\pdfoutput=1

\documentclass[11pt]{article}

\usepackage[]{EMNLP2023}

\usepackage{times}
\usepackage{latexsym}

\usepackage[T1]{fontenc}

\usepackage[utf8]{inputenc}

\usepackage{microtype}

\usepackage{inconsolata}

\usepackage{diagbox}
\usepackage[ruled]{algorithm2e}
\usepackage{amsthm,amsmath,amssymb}
\usepackage{mathrsfs}
\usepackage{multicol}
\usepackage{multirow}
\usepackage{graphicx}
\usepackage{float} 
\usepackage{subfigure}
\usepackage{makecell}
\usepackage{amsfonts}
\usepackage{booktabs}
\title{Using LLM to select the right SQL Query from candidates}

\author{
Zhenwen Li \and Tao Xie \\
School of Computer Science, Peking University, China;\\
  \texttt{\{lizhenwen, taoxie\}@pku.edu.cn}
  }

\begin{document}
\maketitle
\begin{abstract}
Text-to-SQL models can generate a list of candidate SQL queries, and the best query is often in the candidate list, but not at the top of the list.
An effective re-rank method can select the right SQL query from the candidate list and improve the model's performance.
Previous studies on code generation automatically generate test cases and use them to re-rank candidate codes.
However, automatic test case generation for text-to-SQL is an understudied field.
We propose an automatic test case generation method that first generates a database and then uses LLMs to predict the ground truth, which is the expected execution results of the ground truth SQL query on this database.
To reduce the difficulty for LLMs to predict, we conduct experiments to search for ways to generate easy databases for LLMs and design easy-to-understand prompts.
Based on our test case generation method, we propose a re-rank method to select the right SQL query from the candidate list. 
Given a candidate list, our method can generate test cases and re-rank the candidate list according to their pass numbers on these test cases and their generation probabilities.
The experiment results on the validation dataset of Spider show that the performance of some state-of-the-art models can get a 3.6\% improvement after applying our re-rank method.
\end{abstract}

\section{Introduction}
Text-to-SQL is the task of translating a natural language (NL) into a SQL query. 
A text-to-SQL model can generate a candidate list of SQL queries, and sometimes the best query is in the candidate list, but not at the top of the list. For example, the top-10 accuracy of one of the state-of-the-art (SOTA) models has a 7.7\% absolute improvement over top-1 accuracy \cite{re-rank-n-best}.
To better utilize the ability of text-to-SQL models, some studies have focused on selecting the right SQL from the top-K SQL queries in the decoder’s output beam \cite{bogin-etal-2019-global, re-rank-n-best, re-ranker}. 

In code generation, which is a task similar to text-to-SQL, some studies utilize test cases to select the right code from candidate codes, which usually consists of two steps \cite{chen2022codet, shinn2023reflexion}.
First, they use the code generation models to generate assertion statements as test cases. 
Second, they re-rank the candidate codes according to the pass number on test cases and select the best one. 

However, it is much harder for text-to-SQL to get test cases than code generation.
For code generation, the test cases are code assertion statements, which are also code snippets, so they can be generated by the same code generation models.
For text-to-SQL, the input to a text-to-SQL model consists of an NL question and a database. 
A test case for text-to-SQL should contain a test input and an expected execution result.
Test input is the NL question and a new database that has the same schema as the given database, while the expected execution result is the execution results on the new database of the ground truth SQL query.
Because test cases are different from SQL queries, we cannot utilize existing text-to-SQL models to generate them.

To generate test cases for text-to-SQL, the main challenge is to automatically get the expected execution results without the ground truth SQL query, which takes little attention before.
In the study of \citet{testsuiteaccuracy}, they propose a widely used metric \textit{test suite accuracy}. For each NL-SQL pair in the Spider dataset \cite{yu-etal-2018-spider}, they generate a test suite that consists of several databases and the corresponding execution results of the ground truth SQL on these databases. But these test suites need the ground truth SQLs, which is lacking during inference. 
Another study proposes a method to generate small-size databases and lets humans annotate the expected execution results \cite{zhong2022active}. Their goal is to decrease the annotation cost of text-to-SQL data. 
Because their method needs human annotation, it also cannot be used to automatically generate test cases.  

Thanks to the recent improvement in large language models (LLM) such as ChatGPT\footnote{https://chat.openai.com/} and GPT-4 \cite{openai2023gpt4}, we propose a two-step test case generation method for text-to-SQL by leveraging the power of LLMs to predict the expected execution results.
Given a database and the NL question, we first generate a new database that has the same schema as the given database by fuzzing or randomly selecting rows from the given database. We set the maximum table size (MTS) of the generated database to control the size of the database. Then we use a prompt that contains the NL question, the contents of our generated database, and several examples to ask LLMs to predict the expected execution results. We compose our generated database and the expected execution results as a test case.

To improve the prediction accuracy of LLMs, we conduct experiments to explore how to generate easily predicted databases for LLMs and how to design easy-to-understand prompts. 
For database generation, we explore the impact of MTS and the naturalness of database contents on the prediction accuracy of LLMs.
For prompt designing, we explore the impact of the format of database contents and the number of examples in prompts on the prediction accuracy of LLMs.
Based on our experiment results, we choose the optimal hyper-parameters to generate databases and design prompts.

Based on our test case generation method, we propose a three-step re-rank method to select the right SQL query from a candidate list. 
First, we obtain a candidate list from a text-to-SQL model and we classify them according to their execution results on the given databases. 
Second, we generate a test suite that consists of several test cases. For each pair of SQL queries with different classes, there is at least one database that can distinguish them\footnote{Two SQL queries can be distinguished by a database means their execution results are different on this database.}.
Third, we re-rank the candidate list according to their pass number on test cases and their generation probabilities. Then we choose the first SQL query as the output of our re-rank method.

We conduct experiments on the dev dataset of Spider \cite{yu-etal-2018-spider}, the widely used text-to-SQL dataset. We use GPT-4-turbo and GPT-4 to generate test cases, and we follow two state-of-the-art models, DAIL-SQL \cite{DAIL-SQL}, and RESDSQL \cite{resdsql}to generate candidate lists. 
The experimental results show that the performance of DAIL-SQL gets a 3.6\% improvement and RESDSQL gets a 2\% improvement after applying our re-rank methods.

Overall, our study has three main contributions:
\begin{itemize}
    \item We are the first to propose a method to automatically generate test cases for text-to-SQL, without ground truth SQL queries, to the best of our knowledge.

    \item We conduct experiments to explore how to generate easily predicted databases for LLMs and how to design easy-to-understand prompts.

    \item We propose a three-step method to select the right SQL query from a candidate list, and our method can improve the performance of state-of-the-art text-to-SQL models.
\end{itemize}
\section{Test Case Generation}
Our test case generation method consists of two steps: (i) generate a database; and (ii) use LLMs to predict the expected execution results.
\subsection{Database Generation}\label{section.db_generation}
Given a database for text-to-SQL, we use two methods to generate the database.

\noindent\textbf{Fuzzing}. We follow the study of \citet{testsuiteaccuracy} to generate a new database by fuzzing, which is a software testing technique. To maintain the foreign key relation in the new database, we sort the tables of the given database by the foreign key relations before generation. That is, if $tableA$ has a foreign key that refers to a column of $tableB$, $tableB$ will be in front of $tableA$. We then generate the columns of each table in order by randomly generating numbers/strings according to column types.
We use a hyper-parameter maximum table size (MTS) to control the size of each table.
If a column $c_1$ refers to another column $c_2$, we then generate $c_1$ by randomly sampling from $c_2$, rather than randomly generating.  In this way, we can guarantee the foreign key relation between these two columns.

\begin{figure}
\centering 
\subfigure[Two origin tables. The column ``Student ID'' of $Table A$ refers to the column of $Table B$.]{
\label{fig.database_generate.1}
\includegraphics[width=0.45\textwidth]{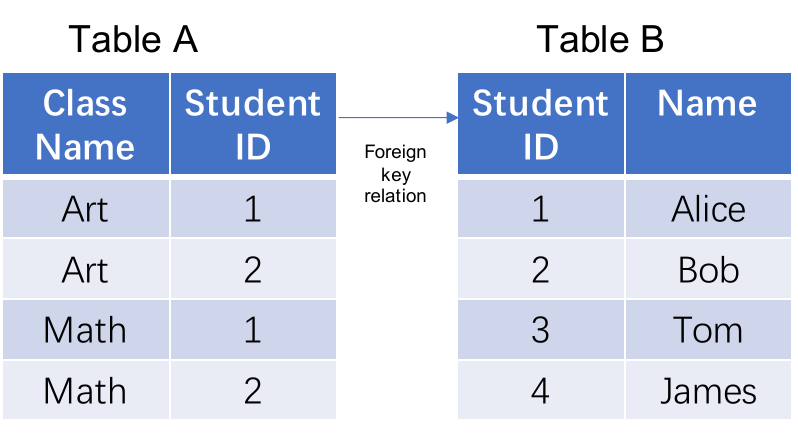}}
\subfigure[An example of generating tables in the wrong order. The new $Table A$ cannot select any rows from the origin $Table A$.]{
\label{fig.database_generate.2}
\includegraphics[width=0.45\textwidth]{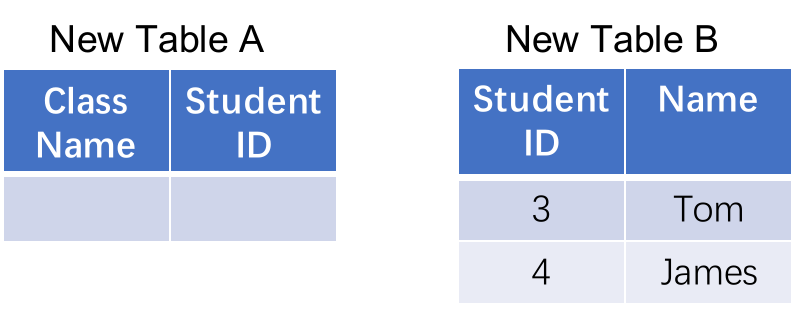}}
\subfigure[An example of generating tables in the right order. The first row of the new $Table B$ is referred to by the new $Table A$, while the second row is randomly selected from the origin $Table B$ (the MTS here is two).]{
\label{fig.database_generate.3}
\includegraphics[width=0.45\textwidth]{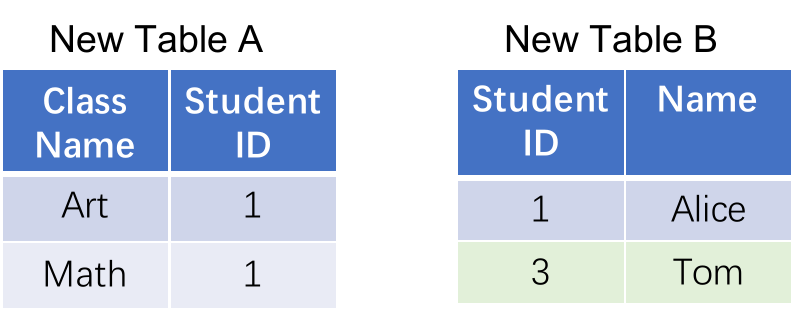}}
\caption{Examples of our random selection algorithm.}
\label{fig:generate_database}
\end{figure}

\noindent\textbf{Random Selection}. Besides fuzzing, We also generate the tables by randomly selecting rows from the origin tables.
Because the grain of our random selection algorithm is row-level, while the grain of fuzzing is cell-level, we cannot generate tables in the same order as fuzzing. For example, as shown in Figure \ref{fig.database_generate.1} and \ref{fig.database_generate.2}, the column ``name id'' of $Table A$ refers to the column of $Table B$. If we generate a new $tableB$ first, there may be no rows in the origin $tableA$ that refer to the new $tableB$. To solve this problem, we generate tables in reverse order. That is, we generate the new $Table A$ before the new $Table B$ by randomly selecting rows from the origin $Table A$.
When we generate the new $Table B$, we first get all the newly generated tables (such as the new $Table A$) that refer to $Table B$ and then collect the referred rows of $Table B$. These rows will be added to the new $Table B$. 
In this way, we can guarantee the foreign key relation between tables.
If there is no referred row or the number of these rows is less than the MTS, we randomly select rows from the origin table to fill it up. 
We show the whole process in Figure \ref{fig.database_generate.3}.
The two methods have advantages and disadvantages. We can get databases with diversity by fuzzing, while random selection can only generate databases with the same rows as the origin database. However, fuzzing may generate databases with unnatural contents, such as people named ``ToT'' or aged ``2333'', while random selection guarantees the naturalness of generated databases. A previous study finds that when people annotate the expected execution results, the accuracy is influenced by the naturalness of the contents \cite{zhong2022active}. For LLMs, we conduct an empirical study to verify these found. 
\subsection{Predict the Expected Execution Result}\label{section.prompt}
\begin{figure}
\centering 
\includegraphics[width=0.45\textwidth]{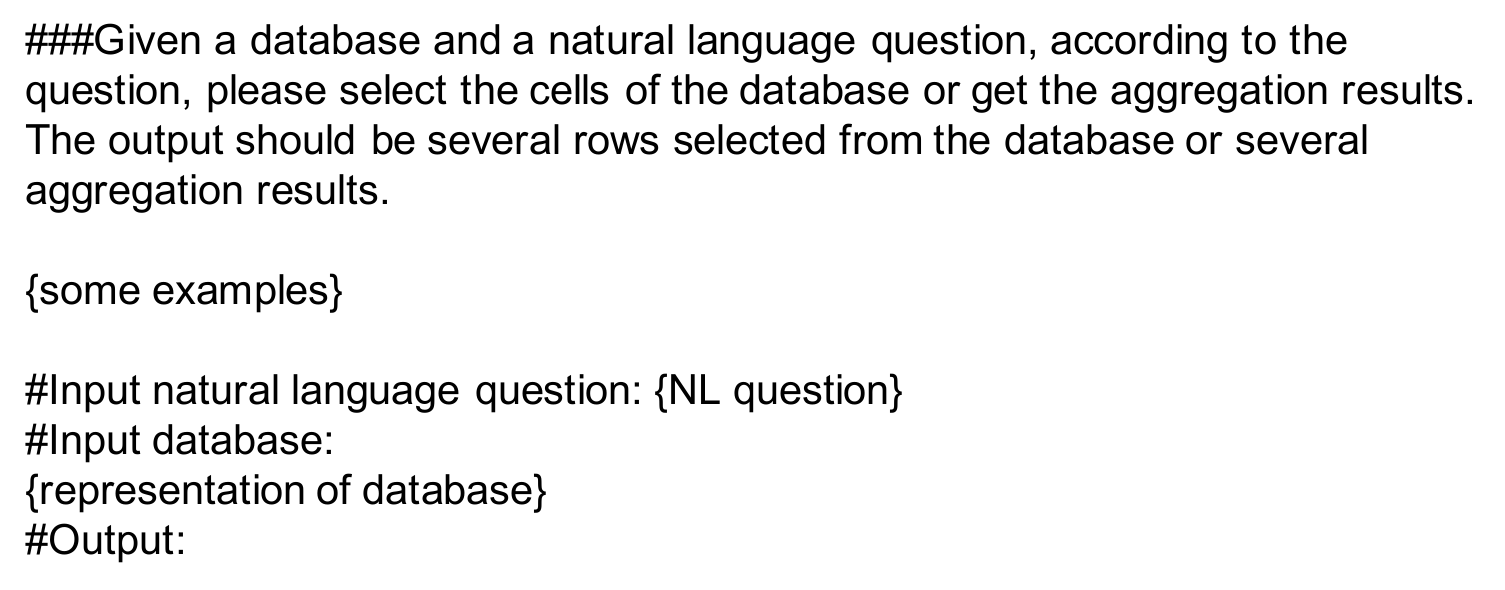}
\caption{The template of our prompt.}
\label{fig:prompt}
\end{figure}
\begin{figure}
\centering 
\includegraphics[width=0.45\textwidth]{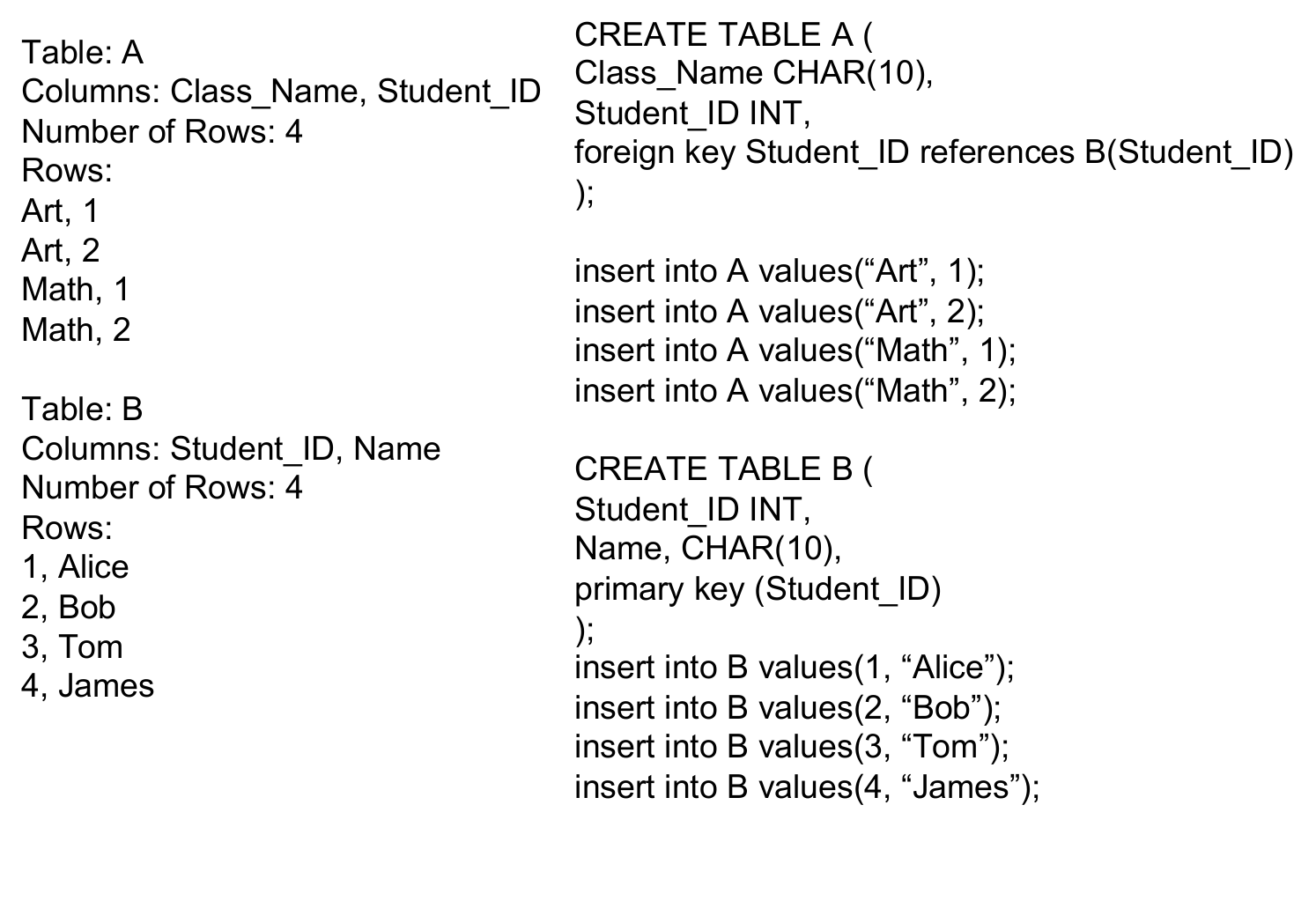}
\caption{Examples of two formats of database representation. The database we use is the above one in Figure \ref{fig.database_generate.1}. The left one is the CSV format and the right one is the SQLite format.}
\label{fig:database_representation}
\end{figure}
After generating a new database, we use LLMs to predict the expected execution result.
As shown in Figure \ref{fig:prompt}, we design a prompt that consists of an instruction, the representation of the database, and the NL questions.
Because the output format is unusual, we need to add some examples to guide the output format and improve the prediction accuracy.

For the representation of the database, we design two formats, the CSV format and the SQLite format. We show examples of these two formats in Figure \ref{fig:database_representation}. The former represents a database by listing the rows, each cell is separated by a comma, just like a CSV file. The latter represents a database by SQLite statements (create table statements and insert value statements).

\section{Candidate Selection}

Given the input of text-to-SQL data, which consists of a database $D$ and an NL question $q$, we propose a three-step method to select the right SQL query from candidate SQL queries by test cases:

\textbf{Step 1.} We use a text-to-SQL model to generate several candidate SQL queries. 
Then we get their execution results on the database $D$.
After classifying candidates according to the execution results, we choose one SQL from each class as $SQL_1, SQL_2,.., SQL_n$.

\begin{algorithm}
    \caption{Test Suite Generation}
    \KwIn{$D\longrightarrow$ Original database}
    \KwIn{$S\longrightarrow$ List of candidate SQL queries}
    \KwIn{$N\longrightarrow$ Threshold of the number of test cases}
    
    $T\leftarrow \{\}$; Store generated test cases.
    
    $C\leftarrow \{\}$; Store classification results of test cases.
    
    \For{$i=$ 1 to $N$}
    {
        $t=generateTestCase(D)$\;
        $c=classify(S, t)$\;
        \If{$c$ not in $C$}
        {
            $T=T\cup t$\;
            $C=C\cup c$\;
            \If{$distinguish(S, T)$}
            {
                return $T$\;
            }
        }
    }
    return $T$\;
    \label{test_suite_generation_algorithm}
\end{algorithm}

\textbf{Step 2.} We generate a test suite consisting of several test cases. 
Algorithm \ref{test_suite_generation_algorithm} describes the overview.
To generate a test suite, we repeatedly generate test cases until these test cases (each one is a <database, expected execution result> pair) can distinguish $SQL_1, SQL_2,.., SQL_n$, or the number of test cases exceeds a threshold.
Each time we generate a new test case, we can classify the candidates into several classes according to the execution results of candidates on the new database.
We will drop the test case if it has the same classification result as one of the existing ones.

\textbf{Step 3.} We compare each execution result of candidate SQL queries with the expected execution result and re-rank these SQL queries according to their pass number on test cases. 
Because sometimes the NL queries are unclear, different columns are selected by candidate SQL queries and LLMs' predictions. 
Taking this into account, we have relaxed our criteria.
Given two execution results, we treat them as the same if their rows are the same and the columns of one are the subset of another.

\section{Experiment}
\subsection{Setting}
\noindent\textbf{Dataset.} We conduct experiments on the well-known text-to-SQL dataset Spider \cite{yu-etal-2018-spider}. Spider is a large-scale cross-domain dataset. It contains 8659 data entries in the training dataset and 1034 in the validation dataset. We use the training dataset to get examples for generating prompts and evaluate our methods on the validation dataset.

\noindent\textbf{Metric.} We use the exact match accuracy (\textbf{EM}) and execution accuracy (\textbf{EX})\footnote{We use the official evaluation scripts at https://github.com/taoyds/test-suite-sql-eval} as the metrics during evaluation.
We also use the code snippets from the official scripts to judge whether two execution results are the same.

\noindent\textbf{LLMs.} We use OpenAI LLMs to conduct experiments. 
We use GPT-3.5-turbo, GPT-4-turbo, and GPT-4 when exploring the optimal hyper-parameters to generate databases and design prompts. 
Because we find GPT-3.5-turbo cannot generate high-quality test cases, we use only  GPT-4-turbo and GPT-4 to re-rank candidate lists.\footnote{\textcolor{red}{The results of GPT-4 are coming soon.}} 

\noindent\textbf{Models.} To generate candidate lists of SQL queries, we use two state-of-the-art models: DAIL-SQL \cite{DAIL-SQL} and RESDSQL \cite{resdsql}. 
For DAIL-SQL, we use the official script and set the hyper-parameters temperature=0.8 and n=20 to generate 20 SQL queries for each data entry of the validation dataset.
For RESDSQL, we also use the official script to generate top-10 SQL queries by beam search.

\noindent\textbf{Bug Report.} During experiments, we find and fix some wrong column types in Spider.
Some string columns should be the type of ``Integer'' / ``Real''.
This can confuse LLMs to predict execution results. 
For example, there is a string column named ``speed''.
It contains two values 100 and 90.
If the NL question asks LLMs to get the maximum speed, the LLMs cannot figure out whether the answer should be the number 100 or the string 90.
\begin{table*}[t!]
  \centering
    \begin{tabular}{l|c|c|c|c|c|c|c}
        \hline
         Model&Base&10 MTS&15 MTS&Fuzzing&SQLite Format&7-shot&9-shot \\
         \hline
         GPT-3.5-turbo&53.6&46.4&43.7&0&55.3&55.0&55.4\\
         GPT-4-turbo&72.0&66.9&64.0&0&71.3&73.2&74.4\\
         GPT-4&78.1&0&0&0&77.2&80.3&80.0\\
        \hline
    \end{tabular}
    \caption{The prediction accuracy of GPT-3.5-turbo and GPT-4 on different hyper-parameters. ``Base'' is our baseline (5 MTS, \textit{Random Selection}, the CSV database format, 5-shot), while the rest are by changing one of the hyper-parameters.}
    \label{tab:hyper-parameter}
\end{table*}
\begin{table}
  \small
  \centering
    \begin{tabular}{lcccc}
        \toprule
        \multirow{2}*{Model}&\multicolumn{2}{c}{Order by}&\multicolumn{2}{c}{Group by}\\
        \cmidrule{2-3}
        \cmidrule{4-5}
         ~&have&not have&have&not have \\
         \midrule
         GPT-3.5-turbo&52.0&53.5&39.4&58.3 \\
         GPT-4-turbo&67.1&73.0&53.7&78.2 \\
         GPT-4&69.6&80.6&63.4&83.4 \\
        \bottomrule
    \end{tabular}
    \caption{The prediction accuracy on data entries having/not having ``group by''/``order by'' clauses.}
    \label{tab:statistic}
\end{table}
\begin{table}
  \centering
    \begin{tabular}{l|c|c}
        \hline
         Model&Order by&Group by \\
         \hline
         GPT-3.5-turbo&52.1&42.1\\
         GPT-4-turbo&67.5&58.8\\
         GPT-4&71.0&68.3\\
        \hline
    \end{tabular}
    \caption{The prediction accuracy after we constrain the range of numbers.}
    \label{tab:constraint_number}
\end{table}
\subsection{Hyper-parameter Optimization}
In this section, we introduce the details of our experiments on exploring how to generate easily predicted databases and easy-to-understand prompts by optimizing the hyper-parameters.
\subsubsection{Hyper-parameters}
Several hyper-parameters influence the difficulty for LLMs to predict the expected execution results.
For database generation, both the MTS and the naturalness of the database influence the difficulty.
For prompt designing, both the format of database content representation and the number of examples influence the difficulty.

\noindent\textbf{MTS}. To explore the influence of database size, during database generation, we control the MTS to get a database of the appropriate size. The MTS we choose are 5, 10, and 15.

\noindent\textbf{Naturalness of database contents}. To explore the influence of the naturalness of database contents, we design two ways to generate a database: \textit{Fuzzing} and \textit{Random Selection}. The former will generate unnatural database contents while the latter select natural contents from the original database. We show the details in Section \ref{section.db_generation}.

\noindent\textbf{Format of database contents}. We design two ways to represent database contents: the CSV format and the SQLite format. We show the details in Section \ref{section.prompt}.

\noindent\textbf{Number of examples}. Because we need examples to guide LLMs to generate a unified output format, we do not conduct zero-shot experiments. The number of examples we choose is 5, 7, and 9.

To reduce the cost, we set default values of hyper-parameters as our baseline and change the hyper-parameter values one by one.
The baseline hyper-parameter we choose is that MTS of 5, natural database contents (\textit{Random Selection}), the \textit{CSV} database contents format, and the number of examples equals 5.
 We choose these values as default because a database with a small size and natural content is easier for humans according to the previous study \cite{zhong2022active}.

To reduce the prompt length, we remove the unused tables and columns.
We parse the ground truth SQL queries in the validation dataset to get the used tables/columns, and then filter the generated database by these tables/columns.
\subsubsection{Results}\label{section:parameter_selection}
We conduct experiments on the validation dataset of Spider. 
We compare the prediction execution results with the actual execution results of the ground truth SQL query.
To eliminate the randomness in database generation, we generate the database three times and compute the average of the prediction accuracy.
We show the experiment results in Table \ref{tab:hyper-parameter}.
As we can see, MTS greatly impacts prediction accuracy. 
Larger MTS lead to larger databases, improving the difficulty for LLMs to predict.
To explore the effect of the naturalness of database contents, we generate databases by \textit{Fuzzing}.
The experiment results show that \textit{Fuzzing} is more difficult to predict than \textit{Random Selection}, which indicates that unnatural database contents can confuse LLMs.
For prompt designing, we generate prompts using the SQLite format. 
The results show it performs worse than the CSV format.
We also change the example numbers in prompts. The results show that 7-shot is the best for GPT-4, while 9-shot is the best for GPT-4-turbo.

According to the experiment results, we can conclude that GPT-4 and GPT-4-turbo can help us automatically generate high-quality test cases, while the prediction accuracy of GPT-3.5-turbo is low. 
To generate high-quality test cases, we should use the \textit{Random Selection} method to generate small-size databases (MTS=5), and the prompt should use SQLite format with 7-9 examples. 
\begin{figure}
\centering 
\includegraphics[width=0.45\textwidth]{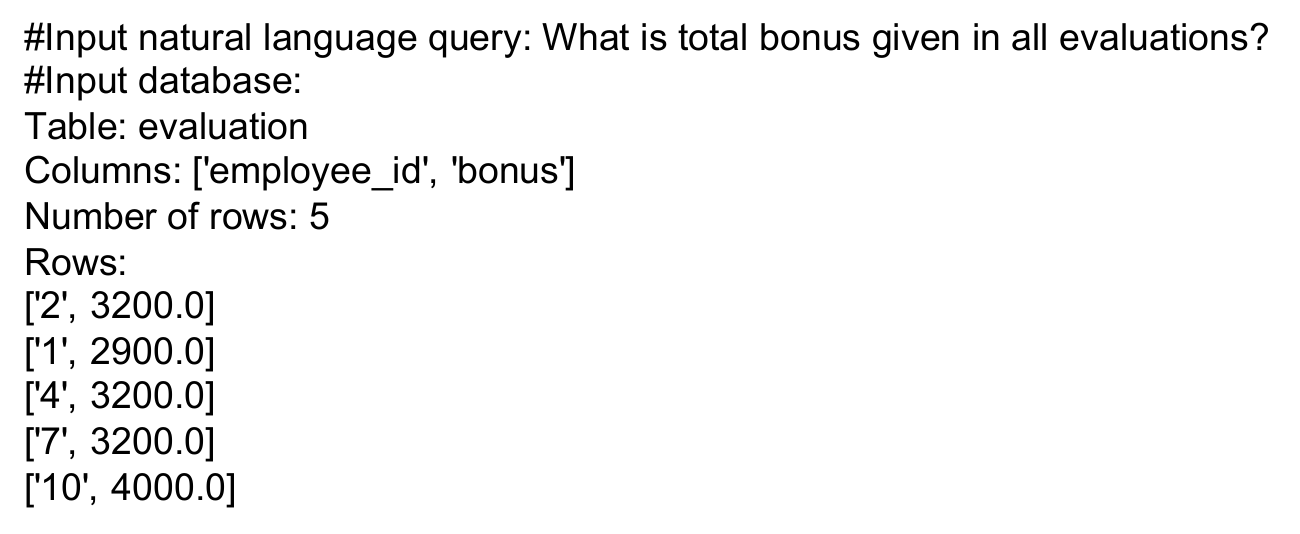}
\caption{An example of our input prompt that asks LLMs to compute the sum of a column.}
\label{fig:large_scale}
\end{figure}
\subsubsection{Constrain the Range of Numbers}
Compared with predicting the SQL query, directly predicting the expected execution results can be more difficult for LLMs when the databases contain large-scale numbers, especially when the NL questions ask LLMs to sort/aggregate some columns. 
We show an example in the Figure \ref{fig:large_scale}. In the example, the LLMs are asked to compute the sum of several large numbers. This is a challenge for the computation power of LLMs.

According to our observation, the NL questions that correspond to SQL queries having ``group by'' clauses usually ask the LLMs to do aggregation operations such as computing the sum/average of columns, while those having ``order by'' clauses usually ask LLMs to sort the columns.
So we select the data entries whose SQL query contains clauses ``group by'' or ``order by'' from the validation dataset, and statistic the prediction accuracy of LLMs.
We show the results in Table \ref{tab:statistic}.
It is challenging for LLMs to predict if the input NL questions correspond to SQL queries containing an ``order by'' or ``group by'' clause, especially for the ``group by'' clause. 
The results indicate that large-scale numbers can confuse LLMs to compute.

To improve the prediction accuracy, we constrain the range of numbers in the columns participating in aggregation/sort operations.
After generating a database, we extract these columns according to the SQL query and replace the numbers as small-scale numbers (the range is an integer between 1-10).
We use the new databases to ask the LLMs to predict execution results, and we show the prediction accuracy in Table \ref{tab:constraint_number}.
The prediction accuracy has been greatly improved, indicating the effectiveness of constraining the range of numbers.
\subsection{Right SQL query Selection}
After selecting the best hyper-parameters for generation test cases, we use test cases to re-rank and select the right SQL query from the candidate lists. We set the hyper-parameter $N=10$ during test suite generation.
Because we cannot use the ground truth SQL query during inference, we parse the SQL queries from candidate lists to get all the used tables/columns and remove the unused tables/columns from prompts.
We also get all the columns participating in the aggregation/sort operations and constrain the range of numbers in them.
To avoid meaningless re-ranking, we re-rank only when at least one of the SQL queries in a candidate list is correct. We also do not re-rank when all the SQL queries in a candidate list are correct.
\subsubsection{Results}
\begin{table}[t!]
  \centering
    \begin{tabular}{l|c|c}
        \hline
         Model&EM&EX\\
         \hline
         DAIL-SQL&63.6&80.9\\
         DAIL-SQL+Self-consistency&64.2&81.8\\
         DAIL-SQL+Re-rank(GPT-4-turbo)&64.5&84.5\\
         DAIL-SQL+Re-rank(GPT-4)&0&0\\
         RESDSQL&76.7&81.9\\
         RESDSQL+Re-rank(GPT-4-turbo)&77.5&83.9\\
         RESDSQL+Re-rank(GPT-4)&0&0\\
        \hline
    \end{tabular}
    \caption{The results of baselines and our method\footnote{Because of the version of GPT-4, the results of the DAIL-SQL are slightly different from their paper's report.}.}
    \label{tab:main_result}
\end{table}
We show the results in Table \ref{tab:main_result}. 
Our re-rank method can significantly improve both EM and EX accuracy.
Our method improves the EM accuracy little because the re-rank is based on the execution results of SQL queries.
DAIL-SQL + Self-consistency is using voting to select the right SQL query.
Our re-rank method performs better than DAIL-SQL + Self-consistency, indicating that re-ranking based on test cases is better than voting.

\begin{figure*}
\centering 
\includegraphics[width=1\textwidth]{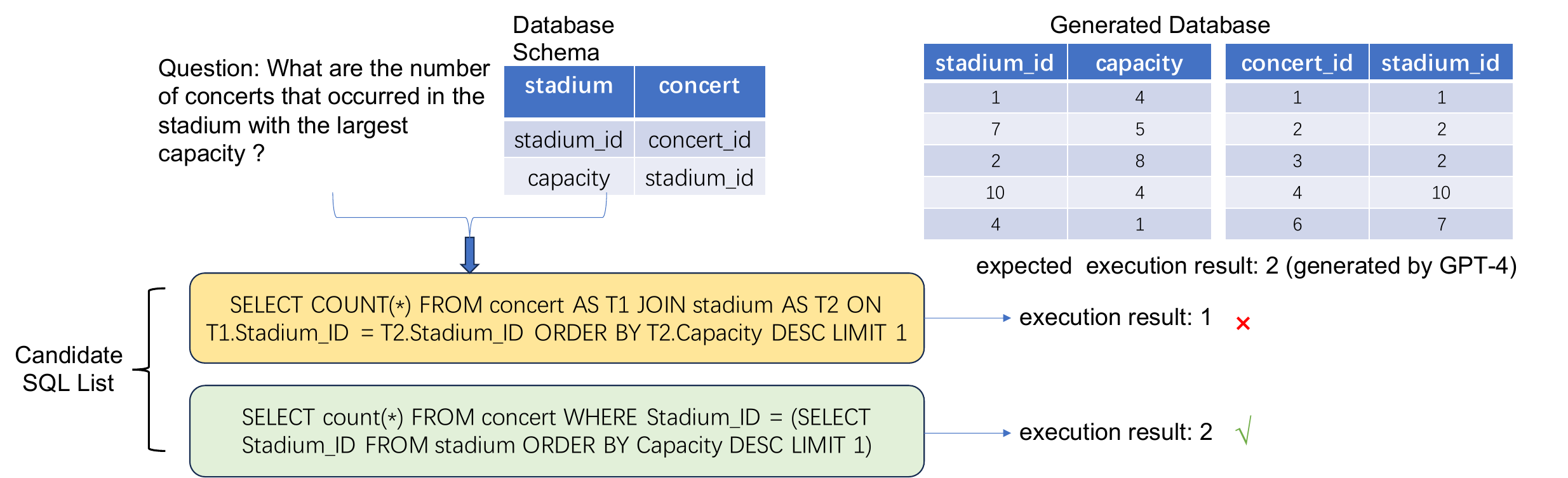}
\caption{An example of our re-rank method.}
\label{fig:case_study}
\end{figure*}

For DAIL-SQL, 201 candidate lists are re-ranked, because we do not re-rank the candidate list when there is no correct one or all of the SQL queries are the same. 
For RESD-SQL, 659 candidate lists are re-ranked.
We statistic the prediction accuracy on test cases for these candidate lists.
The GPT-4-turbo achieves a prediction accuracy of 70.1\%, while the GPT-4 achieves a prediction accuracy of xx\%.
The prediction accuracy is lower than that of the experiments in Section \ref{section:parameter_selection}. 
This is because the re-ranked candidate lists usually correspond to difficult data entries, while the results in Section \ref{section:parameter_selection} come from all the data entries in the validation dataset.
Even if the prediction accuracy is far from perfect, these test cases can still help re-rank the candidate lists and significantly improve the performance of the original model.
\subsubsection{Case Study}
We show an example of our re-rank method in Figure \ref{fig:case_study}.
In this case, the DAIL-SQL generates several SQL queries.
After classifying according to their execution results, these SQL queries are classified into two groups.
We show a SQL query of each group in the left part of the figure.
Our test case generation method automatically generates a test case, consisting of a database and an expected execution result which is generated by GPT-4.
The generated database contains three tables, each having several columns.
We show in the figure only the tables/columns that appear in one of the candidate SQL queries.
When we design the prompt to ask for expected execution results, we also represent the database with only these tables/columns.
The ``capacity'' column participates in sort operation in both the candidate SQL queries.
So we replace the numbers in this column with small-range integer numbers (range between 1-10).

In this case, the database content is natural and the representation of the database is small (only two tables, each containing two columns and five rows).
So it is easy for LLMs to predict and the GPT-4 predicts the right expected execution result. 
After comparing the execution result of each SQL query with the expected one, we select the SQL query that passes this test case.
\section{Related Work}
In this section, we introduce the applications of LLM in text-to-SQL, the relationship between our work and previous re-ranking studies, and discuss the advantages of our database generation algorithm over previous work. 
\subsection{Text-to-SQL with LLM}
To utilize the ability in natural language understanding (NLU) of LLMs and their world knowledge, previous work \cite{resdsql, scholak2021picard, qi2022rasat-sql, rubin2021smbop} encodes both the input NL utterances and database structures (table/column names) by LLMs such as BERT, RoBERTa, and BART \cite{lewis-etal-2020-bart}. 
With the improvement in natural language generation (NLG) LLM such as Codex \cite{codex}, ChatGPT, and GPT-4, there are some prompting approaches using LLM on text-to-SQL. DIN-SQL \cite{pourreza2023dinsql}decomposes text-to-SQL into several sub-tasks, and design prompts for each sub-task. 
One of the state-of-the-art models for Spider
DAIL-SQL \cite{DAIL-SQL} uses an example selection method to select examples from the training dataset according to question similarity between training data entries and validation/test data entries.
Then it uses few-shot prompts that consist of these selected examples to ask GPT-4 to generate SQL queries.
In the work of \citet{rajkumar2022evaluatingsql}, they evaluate the performance of Codex and GPT-3 \cite{gpt3} on Spider. In the work of \citet{gptfortext2sql}, they evaluate the zero-shot performance of ChatGPT on text-to-SQL task.
These prompting approaches ask LLMs to directly generate SQL queries.
Compared with them, our method uses LLMs to generate test cases for selecting the right SQL query.
So our method can be seen as a post-process for these methods. 
\subsection{Re-ranking}
Re-ranking is widely used in many deep learning tasks \cite{related-work-rerank1, related-work-rerank2, related-work-rerank3}.
In text-to-SQL, some work designs a re-ranker that can predict the generation probability of a SQL \cite{re-ranker}, and some train a model to re-rank candidates based on the global alignment of database constants to question words \cite{bogin-etal-2019-global}.
In the work of \citet{re-rank-n-best}, they build a multi-label classification model to predict the class of the SQL queries.
But no work uses test cases to re-rank SQL queries, to our best knowledge. 

In the work of \citet{chen2022codet}, to re-rank the candidates from the code generation task, they generate some test cases for these candidate codes, and then execute the candidate codes on these test cases. However, generating test cases of text-to-SQL is much harder than that of code generation, so using test cases to re-rank SQL queries is still understudied.
\subsection{Database Generation}
To distinguish different SQL queries, previous work uses fuzzing to generate databases and proposes a new metric named test suite accuracy \cite{testsuiteaccuracy}. However, the generated databases are too large for LLMs, so they cannot be used to generate test cases.
Some studies generate small-size databases to distinguish SQL queries \cite{databasegeneration, zhong2022active}.
In the work of \citet{zhong2022active}, they propose a two-stage database generation algorithm that first generates a large database and then randomly drops the records. The goal of their work is to decrease the cost of annotating text-to-SQL data.
Their goal is to generate databases to distinguish as many candidate SQL queries as possible to reduce the cost of human annotation.
Compared with them, our method uses LLMs to predict the expected execution results.
The cost of LLMs is little but the prediction accuracy is lower than humans. 
So we need to generate more databases than them.
Some of our databases have the same classification results for candidate lists.
This redundancy is important for improving the performance of our re-rank method.
\section{Conclusion}
In this paper, we use test cases to re-rank candidate lists, which are generated by text-to-SQL models, and select the right SQL query.
To obtain test cases, we propose an automatic test case generation method that generates databases and uses LLMs to predict the expected execution results.
We conduct experiments to explore how to generate databases for LLMs to easily predict and how to design prompts to utilize the power of LLMs.
We conduct experiments on the Spider dataset to re-rank candidate lists.
The experiment results show the effectiveness of our re-rank method.

Our study shows that using test cases to re-rank candidate lists can largely improve the performance of text-to-SQL models.
This research direction holds significant value and potential.
Future studies can explore how to generate higher quality test cases and how to better explore these test cases.
\section{Limitation}
Our work mainly contains three limitations.
First, our database generation method for test cases has limitations. It only selects rows from the original databases, so that cannot distinguish some SQL queries. 
Second,  only about 60\% of our test cases are correct, reducing the final EM/EX accuracy.
Third, our re-rank method costs time and tokens to select the right SQL query, because we use an average of ten times OpenAI's API for generating test cases.

\bibliography{anthology,custom}
\bibliographystyle{acl_natbib}

\end{document}